\documentclass[11pt,a4paper]{article}
\usepackage{amsmath}
\usepackage{naaclhlt2018}
\usepackage{times}
\usepackage{latexsym}
\usepackage{paralist}
\usepackage{url}
\usepackage{comment}
\usepackage{amssymb} 
\usepackage{amsfonts}
\usepackage{subcaption}
\RequirePackage{bm}

\usepackage{todonotes}
\newcommand{\Note}[2]{} 
\newcommand{\SideNote}[2]{} 
\renewcommand{\Note}[2]{\todo[color=#1,size=\small, inline=true]{#2}} 
\renewcommand{\SideNote}[2]{\todo[color=#1,size=\small]{#2}} %

\usepackage{dsfont}
\usepackage{graphicx,color}  
\graphicspath{ {figures/} }

\usepackage{makecell}

\aclfinalcopy 

\title{Learning Word Embeddings for Low-resource Languages by PU Learning}

\author{
Chao Jiang\\
University of Virginia \\
 {\tt cj7an@virginia.edu} \\ 
 \\
 {\bf Cho-Jui Hsieh} \\
University of California Davis \\
 {\tt chohsieh@ucdavis.edu} \\
\And
Hsiang-Fu Yu\\
Amazon \\
 {\tt rofuyu@cs.utexas.edu} \\ 
 \\
{\bf Kai-Wei Chang}\\
University of California Los Angeles \\
 {\tt kwchang@cs.ucla.edu}
}

\date{}

\begin{document}
\maketitle

\begin{abstract}
Word embedding is a key component in many downstream applications in processing natural languages. Existing approaches often assume the existence of a large collection of text for learning effective word embedding. However, such a corpus may not be available for some low-resource languages. In this paper, we study how to effectively learn a word embedding model on a corpus with only a few million tokens. In such a situation, the co-occurrence matrix is  sparse as the co-occurrences of many word pairs are unobserved. In contrast to existing approaches often only sample a few unobserved word pairs as negative samples, we argue that the zero entries in the co-occurrence matrix also provide valuable information. We then design a Positive-Unlabeled Learning (PU-Learning) approach to factorize the co-occurrence matrix and validate the proposed approaches in four different languages. 


\end{abstract}

\section{Introduction}

Learning word representations has become a fundamental problem in processing natural languages. 
These semantic representations, which map a word into a point in a linear space, have been widely applied in downstream applications, including named entity
recognition ~\cite{guo2014}, document ranking ~\cite{NMCC16}, sentiment analysis~\cite{IrsoyCardie14}, question answering~\cite{VQA}, and image captioning~\cite{karpathy2015deep}.

Over the past few years, various approaches have been proposed to learn word vectors~(e.g., ~\cite{pennington2014glove,Mikolov2013,OL14a,ji2015wordrank}) based on co-occurrence information between words observed on the training corpus. The intuition behind this is to represent words with similar vectors if they have similar contexts. To learn a good word embedding, most approaches assume a large collection of text is freely available, such that the estimation of word co-occurrences is accurate. For example, the Google Word2Vec model~\cite{Mikolov2013}  is trained on the Google News dataset, which contains around 100 billion tokens, and the GloVe embedding~\cite{pennington2014glove} is trained on a crawled corpus that contains 840 billion tokens in total.
However, such an assumption may not hold for low-resource languages such as Inuit or Sindhi, which are not spoken by many people or have not been put into a digital format. 
For those languages, usually, only a limited size corpus is available. Training word vectors under such a setting is a challenging problem.


One key restriction of the existing approaches is that they often mainly rely on the word pairs that are observed to co-occur on the training data. When the size of the text corpus is small, most word pairs are unobserved, resulting in an extremely sparse  co-occurrence matrix (i.e., most entries are zero)\footnote{Note that the zero term can mean either the pairs of words cannot co-occur or the co-occurrence is not observed in the training corpus.}. For example, the text8\footnote{http://mattmahoney.net/dc/text8.zip} corpus has about 17,000,000 tokens and 71,000 distinct words. The corresponding co-occurrence matrix has more than five billion entries, but only about 45,000,000 are non-zeros (observed on the training corpus). Most existing approaches, such as Glove and Skip-gram, cannot handle a vast number of zero terms in the co-occurrence matrix; therefore, they only sub-sample a small subset of zero entries during the training.

In contrast, we argue that the {\it unobserved} word pairs can provide valuable information for training a word embedding model, especially when the co-occurrence matrix is very sparse. Inspired by the success of Positive-Unlabeled Learning (PU-Learning) in collaborative filtering applications~\cite{RP08a-short,YH08a,RP09a,QT10a,UP13a-short,hsieh2015pu}, we design an algorithm to effectively learn  word embeddings from both positive (observed terms) and unlabeled (unobserved/zero terms) examples. 
Essentially, by using the square loss to model the unobserved terms and designing an efficient update rule based on linear algebra operations, the proposed PU-Learning framework can be trained efficiently and effectively.


We evaluate the performance of the proposed approach in English\footnote{Although English is not a resource-scarce language, we simulate the low-resource setting in an English corpus. In this way, we leverage the existing evaluation methods to evaluate the proposed approach.} and  other three resource-scarce languages. We collected unlabeled language corpora from Wikipedia and compared the proposed approach with popular approaches, the Glove and the Skip-gram models, for training word embeddings. The experimental results show that our approach significantly outperforms the baseline models, especially  when the size of the training corpus is small.

Our key contributions are summarized below.
\begin{itemize}
    \item We propose a PU-Learning framework for learning word embedding.
    \item We tailor the coordinate descent algorithm~\cite{Yu} for solving the corresponding optimization problem. 
    \item Our experimental results show that PU-Learning improves the word embedding training in the low-resource setting.
\end{itemize}

\section{Related work}

\paragraph{Learning word vectors.}
The idea of learning word representations can be traced back to Latent Semantic Analysis (LSA)~\cite{deerwester1990indexing} and Hyperspace Analogue to Language (HAL)~\cite{lund1996producing}, where word vectors are generated by factorizing a word-document and word-word co-occurrence matrix, respectively. 
Similar approaches can also be extended to learn other types of relations between words~\cite{yih2012polarity,chang2013multi} or entities~\cite{CYYM14}.
However, due to the limitation of the use of principal component analysis, these approaches are often less flexible. Besides, directly factorizing the co-occurrence matrix may cause the frequent words dominating the training objective.

In the past decade, various approaches have been proposed to improve the training of word embeddings. For example, instead of factorizing the co-occurrence count matrix, \newcite{bullinaria2007extracting,OL14a} proposed to factorize 
\textit{point-wise mutual information} (PMI)  and \textit{positive PMI} (PPMI) matrices as these metrics scale the co-occurrence counts~\cite{bullinaria2007extracting,OL14a}. Skip-gram model with negative-sampling (SGNS) and  Continuous Bag-of-Words models~\cite{Mikolov2013SGNS} were proposed for training word vectors on a large scale without consuming a large amount of memory. GloVe~\cite{pennington2014glove} is proposed as an alternative to decompose a weighted log co-occurrence matrix with a bias term added to each word. Very recently, \textit{WordRank} model ~\cite{ji2015wordrank} has been proposed  to minimize a ranking loss which naturally fits the tasks requiring ranking based evaluation metrics. \newcite{stratos2015model} also proposed CCA (canonical correlation analysis)-based word embedding which shows competitive performance. All these approaches focus on the situations where a large text corpus is available.

\paragraph{Positive and Unlabeled (PU) Learning:} 
Positive and Unlabeled (PU) learning~\cite{li2005learning} is proposed for training a model when the positive instances are partially labeled and the unlabeled instances are mostly negative. Recently, PU learning has been used in many classification and collaborative filtering applications due to the nature of ``implicit feedback'' in many recommendation systems---users usually only provide positive feedback (e.g., purchases, clicks) and it is very hard to collect negative feedback.


To resolve this problem, a series of PU matrix completion algorithms have been proposed~\cite{RP08a-short,YH08a,RP09a,QT10a,UP13a-short,hsieh2015pu,Yu}. The main idea is to assign a  small uniform weight to all the
missing or zero entries and factorize the corresponding matrix. Among them, \newcite{Yu} proposed an efficient algorithm for matrix factorization with PU-learning, such that the weighted matrix is constructed implicitly. In this paper, we design a new approach for training word vectors by leveraging the PU-Learning framework and existing word embedding techniques. To the best of our knowledge, this is the first work to train word embedding models using the PU-learning framework.

\section{PU-Learning for  Word Embedding}
\label{sec:algo}

Similar to GloVe and other word embedding learning algorithms, the proposed approach consists of three steps. The first step is to construct a co-occurrence matrix. Follow the literature~\cite{levy2014linguistic}, we use the PPMI metric to measure the co-occurrence between words. Then, in the second step, a PU-Learning approach is applied to factorize the co-occurrence matrix and generate word vectors and context vectors. Finally, a post-processing step generates the final embedding vector for each word by combining the word vector and the context vector.

We summarize the notations used in this paper in Table~\ref{table:notation} and describe the details of each step in the remainder of this section.

\begin{table}[t]
\centering
\label{table:notation}
\begin{tabular} {l|l}

$\mathcal{W}, \mathcal{C}$ & vocabulary of central and context words \\
$m, n$ & vocabulary sizes   \\ 
 $k$ & dimension of word vectors \\
 $W, H$ & $m \times k$ and $n \times k$ latent matrices \\
$C_{ij}$ & weight for the (i, j) entry \\
$A_{ij}$ & value of the  PPMI matrix \\
 $Q_{ij}$ & value of  the co-occurrence matrix \\
 $\bw_i, \bh_j$ &  $i$-th row of $W$ and $j$-th row of $H$ \\
 $\bb, \hat{\bb}$ &  bias term  \\
$\lambda_i, \overline{\lambda}_j $ & regularization parameters \\
 $|\cdot|$ & the size of a set \\
 $\Omega$ & Set of possible word-context pairs \\
 $\Omega^+ $ & Set of observed word-context pairs  \\
 $\Omega^-$ & Set of unobserved word-context pairs  \\
\end{tabular}
\caption{Notations.}
\label{table:notation}

\end{table}


\subsection{Building the Co-Occurrence Matrix}
\label{sec:building-pmi}

Various metrics can be used for estimating the co-occurrence between words in a corpus. 
PPMI metric stems from \textit{point-wise mutual information} (PMI) which has been widely used as a measure of word association in NLP for various tasks \cite{church1990word}. 
In our case, each entry $PMI(w, c)$ represents the relevant measure between a word $w$ and a context word $c$
by calculating the ratio between their joint probability (the chance they appear together in a local context window) and their marginal probabilities (the chance they appear independently)~\cite{OL14a}. More specifically, each entry of PMI matrix can be defined by
\begin{equation}
     PMI(w, c) = \log \frac{\hat{P}(w, c)}{\hat{P}(w) \cdot \hat{P}(c)},  
\end{equation}
where $\hat{P}(w), \hat{P}(c)$ and $\hat{P}(w, c)$ are the the frequency of word $w$, word $c$, and word pairs $(w, c)$, respectively. The PMI matrix can be computed based on the co-occurrence counts of word pairs, and it is an information-theoretic association measure which effectively eliminates the big differences in magnitude among entries in the co-occurrence matrix.

Extending from the PMI metric, the PPMI metric replaces all the negative entries in PMI matrix by 0:
\begin{equation}
    PPMI (w, c) = \max(PMI(w, c), 0).
\end{equation}
The intuition behind this is that people usually perceive  positive associations between words (e.g. ``ice'' and ``snow''). In contrast, the negative association is hard to define~\cite{OL14a}. Therefore, it is reasonable to replace the negative entries in the PMI matrix by 0, such that the negative association is treated as ``uninformative''.
Empirically, several existing works~\cite{Levy2015,bullinaria2007extracting} showed that the PPMI metric achieves good performance on various semantic similarity tasks.

%

In practice,  we follow the pipeline described in ~\newcite{Levy2015} to build the PPMI matrix and apply several useful tricks to improve its quality. First, we apply a context distribution smoothing mechanism to enlarge the probability
of sampling a rare context. In particular, all context counts are scaled to the power of $\alpha$.\footnote{Empirically, $\alpha = 0.75$ works well ~\cite{Mikolov2013SGNS}.}:
\begin{equation*}
\label{pmi_function}
 PPMI_\alpha(w, c) = \max \left( \log \frac{\hat{P}(w, c)}{\hat{P}(w)\hat{P}_\alpha(c)}, 0\right) 
 \end{equation*}
\begin{equation*}
\hat{P}_\alpha(c) = \frac{\#(c)^\alpha}{\sum_{\bar c}\#(\bar c)^\alpha},
\end{equation*}
where $\#(w)$ denotes the number of times word $w$ appears.
This smoothing mechanism effectively alleviates PPMI's bias towards rare words ~\cite{Levy2015}. 

Next, previous studies show that words that occur too frequent often dominate the training objective~\cite{Levy2015} and degrade the performance of word embedding. To avoid this issue, we follow \newcite{Levy2015} to sub-sample words with frequency more than a threshold $t$ with a probability $p$ defined as:
\begin{equation*}
p = 1 - \sqrt{\frac{t}{\hat{P}(w)}}.
\end{equation*}

\subsection{PU-Learning for Matrix Factorization }

We proposed a matrix factorization based word embedding model which aims to minimize the reconstruction error on the PPMI matrix. 
The low-rank embeddings are obtained by solving the following optimization problem: 
\begin{align} 
\min_{W, H}&\sum_{i, j \in \Omega} C_{ij}(A_{ij} - \bw_i^T \bh_j - b_i - \hat{b}_j)^2 \nonumber\\
&+ \sum_i \lambda_i \|\bw_i\|^2 + \sum_j \overline{\lambda}_j\|\bh_j\|^2, 
\label{eq:obj}
\end{align}
where $W$ and $H$ are   $m \times k$ and $n \times k$ latent matrices, representing words and context words, respectively. 
The first term in Eq. \eqref{eq:obj} aims for minimizing reconstruction error, and the second and third terms are regularization terms. 
$\lambda_i$ and $\overline{\lambda}_j$ are weights of regularization term. They are hyper-parameters that need to be tuned.


The zero entries in co-occurrence matrix denote that two words never appear together in the current corpus, which also refers to unobserved terms. 
The unobserved term can be either real zero  (two words shouldn't be co-occurred even when we use very large corpus)
or just missing in the small corpus. 
In contrast to SGNS sub-sampling a small set of zero entries as negative samples, our model will try to use the information from all zeros.

The set $\Omega$ includes all the $|\mathcal{W}|\times |\mathcal{C}|$ entries---both positive and zero entries:
\begin{equation}
\Omega = \Omega^+ \cup \Omega^-.
\end{equation}
Note that we define the positive samples $\Omega^+$ to be all the $(w,c)$ pairs that appear at least one time in the corpus, 
and negative samples $\Omega^-$ are word pairs that never appear in the corpus.

\paragraph{Weighting function.} 
Eq~\eqref{eq:obj} is very similar to the one used in previous matrix factorization approaches such as GloVe, but
we propose a new way to set the weights $C_{ij}$. 
If we set equal weights for all the entries, then $C_{ij}=\text{constant}$, and the model is very similar to conducting
SVD for the PPMI matrix. Previous work has shown that this approach often suffers from poor performance~\cite{pennington2014glove}. 
More advanced methods, such as GloVe, set non-uniform weights for observed entries to reflect their confidence. 
However, the time complexity of their algorithm is proportional to number of nonzero weights ($|(i,j)\mid C_{ij}\neq 0|$), 
thus they have to set zero weights for all the unobserved entries
 ($C_{ij}=0$ for $\Omega^-$), or try to incorporate a small set of unobserved entries by negative sampling. 
 
We propose to set the weights for $\Omega^+$ and $\Omega^-$ differently using the following scheme: 
\begin{equation}
\begin{split}
    &C_{ij}= \\
&\begin{cases}
    (Q_{ij}/x_{max})^\alpha,& \text{if } Q_{ij}\leq x_{max} \text{, and $(i, j)\in\Omega^+$ }\\
    1,& \text{if } Q_{ij}> x_{max} \text{, and $(i,j)\in\Omega^+$}\\
    \rho,              & \text{$(i,j)\in\Omega^-$}
\end{cases}
\end{split}
\label{weighting_function}
\end{equation}
Here $x_{max}$  and $\alpha$ are re-weighting parameters, and $\rho$ is the unified weight for unobserved terms. We will discuss them later.



For entries in $\Omega^+$, we set the non-uniform weights as in GloVe~\cite{pennington2014glove}, which assigns larger weights to context word that appears more often with the given word, but also avoids overwhelming the other terms. For entries in $\Omega^-$, 
instead of setting their weights to be 0, we assign a small constant weight $\rho$. The main idea is from the literature of PU-learning \cite{YH08a,hsieh2015pu}: although missing
entries are highly uncertain, they are still likely to be true 0, so we should incorporate them in the learning process
but multiplying with a smaller weight according to the uncertainty. Therefore, $\rho$ in \eqref{weighting_function} reflects how confident we are to the zero entries. 

In our experiments, we set $x_{max} = 10, \alpha = 3/4$ according to ~\cite{pennington2014glove}, and let $\rho$ be a parameter to tune.
Experiments show that adding weighting function obviously improves the performance especially on analogy tasks.

\paragraph{Bias term.} Unlike previous work on PU matrix completion~\cite{Yu,hsieh2015pu}, 
we add the bias terms for word and context word vectors. Instead of directly using $\bw^\top_i \bh_j$ to approximate $A_{ij}$, we use $$A_{ij} \approx \bw^\top_i \bh_j + b_i + \hat{b}_j. $$

\newcite{Yu} design an efficient column-wise coordinate descent algorithm for solving the PU matrix factorization
problem; however, they do not consider the bias term in their implementations. To incorporate the bias term in \eqref{eq:obj}, we propose the following training algorithm based on the coordinate descent approach. Our algorithm does not introduce much overhead compared to that in \cite{Yu}. 

We augment each $\bw_i, \bh_j\in R^k$ into the following $(k+2)$ dimensional vectors: 
\begin{center}
$\bw_i' = \left[  
  \begin{array}{c}  
          w_{i1} \\  
          \vdots \\  
          w_{ik}  \\
          1 \\
          b_i
 \end{array}  
 \right] $   \quad   $\bh_j' = \left[  
  \begin{array}{c}  
          h_{j1} \\  
          \vdots \\  
          h_{jk}  \\
          \hat{b}_j \\
          1
 \end{array}  
 \right]$
 \end{center}

Therefore, for each word and context vector, we have the following equality
 $$\langle \bw_i', \bh_j'\rangle = \langle \bw_i, \bh_j\rangle + b_i + \hat{b}_j, $$
 which means the loss function in \eqref{eq:obj} can be written as
 \begin{align*} 
\sum_{i, j \in \Omega} C_{ij}(A_{ij} - \bw_i'^\top \bh_j')^2.
\end{align*}
Also, we denote $W'=[\bw_1', \bw_2', \dots, \bw_n']^\top$ and $H'=[\bh_1', \bh_2', \dots, \bh_n']^\top$. 
In the column-wise coordinate descent method, at each iteration we pick a $t\in\{1, \dots, (k+2)\}$, and update
the $t$-th column of $W'$ and $H'$. The updates can be derived for the following two cases: 
\begin{itemize}
\item[a.] When $t\leq k$, the elements in the $t$-th column is $w_{1t}, \dots, w_{nt}$ and
we can directly use the update rule derived in~\newcite{Yu} to update them. 
\item[b.]  When $t=k+1$, we do not update the corresponding
column of $W'$ since the elements are all 1, and we use the similar coordinate descent update to update the $k+1$-th column
of $H'$ (corresponding to $\hat{b}_1, \dots, \hat{b}_n$). When $t=k+2$, we do not update the corresponding
column of $H'$ (they are all 1) and we update the $k+2$-th column of $W'$ (corresponding to $b_1, \dots, b_n$) using coordinate descent. 
\end{itemize}
\vspace{10pt}
With some further derivations, we can show that the algorithm only requires $O(\text{nnz}(A) + nk)$ time to update
each column,\footnote{Here we assume $m=n$ for the sake of simplicity. And, nnz(A) denotes the number of nonzero terms in the matrix A.} so the overall complexity is $O(\text{nnz}(A)k + nk^2)$ time per epoch, which is only proportional to number of nonzero terms in $A$. Therefore, with the same time complexity as GloVe, we can utilize the information from all the 
zero entries in $A$ instead of only sub-sampling a small set of zero entries.


\subsection{Interpretation of Parameters}
\label{sec:rho_and_lambda}

In the PU-Learning formulation, $\rho$ represents the unified weight that assigned to the unobserved terms. Intuitively, 
$\rho$ reflects the confidence on unobserved entries---larger $\rho$ means that we are quite certain about the zeroes, 
while small $\rho$ indicates the  many of unobserved pairs are  not truly zero. When $\rho=0$, 
the PU-Learning approach reduces to a model similar to GloVe, which discards all the unobserved terms. 
In practice, $\rho$ is an important  parameter to tune, and we find that $\rho = 0.0625$ achieves the best results in general.
Regarding the other parameter, $\lambda$ is the regularization term for preventing the embedding model from over-fitting. In practice, we found the performance
is not very sensitive to $\lambda$ as long as it is reasonably small. More discussion about the parameter setting can be found in Section \ref{sec:exp-para}.
%
%

\paragraph{Post-processing of Word/Context Vectors}
\label{sec:post-process-embedding}
The PU-Learning framework factorizes the PPMI matrix and generates two vectors for each word $i$, $\bw_i \in R^k$ and $\bh_i\in R^k$. The former represents the word when it is the central word and the latter represents the word when it is in context.
\newcite{Levy2015} shows that averaging these two vectors ($\bu_i^{\text{avg}} = \bw_i + \bh_i$)  leads to consistently better performance. The same trick of constructing word vectors is also used in GloVe. Therefore, in the experiments, we evaluate all models with $\bu^{\text{avg}}$.

\section{Experimental Setup}
\label{sec:exp}


Our goal in this paper is to train word embedding models for low-resource languages. In this section, we describe the experimental designs to evaluate the proposed PU-learning approach. We first describe the data sets  and the evaluation metrics. Then, we provide details of parameter tuning.

\begin{table*}[h]
\centering
\begin{tabular}{c||c|c|c|c|c||c|c}

\hline
 & \multicolumn{5}{c||}{Similarity task} & \multicolumn{2}{c}{Analogy task} \\ 
 \hline
Word embedding & WS353 &  Similarity &  Relatedness & M. Turk  & MEN & 3CosAdd & 3CosMul \\ 
GloVe & 48.7 & 50.9 & 53.7 & 54.1 & 17.6 & 32.1 &	28.5 \\ 
SGNS & 67.2	 & 70.3	& 67.9 & $~\textbf{59.9}^{*}$ & $~\textbf{25.1}^{*}$  & 30.4 &	27.8 \\   	
\hline
PU-learning & $~\textbf{68.3}^{*}$ & $~\textbf{71.8}^{*}$ & $~\textbf{68.2}^{*}$ & 57.0 & 22.7 & $~\textbf{32.6}^{*}$ & $~\textbf{32.3}^{*}$ \\
\hline

\end{tabular}
\caption{Performance of the best SGNS, GloVe, PU-Learning models, trained on the text8 corpus. Results show that
 our proposed model is better than SGNS and GloVe. Star indicates it is significantly better than  the second best algorithm in the same column according to Wilcoxon signed-rank test. ($p < 0.05$)}
\label{table:comparingontext8}

\end{table*}

\begin{table*}[h]
\centering

\label{table:comparing}
\begin{tabular}{c||c|c|c|c|c||c}
\hline
 & \multicolumn{5}{c||}{Similarity task} & \multicolumn{1}{c}{Analogy task} \\ 
 \hline
Language & WS353 &  Similarity &  Relatedness & M. Turk  & MEN & Google \\ 
\hline
English (en) & 	353 &	203 &	252 &	287 & 3,000 &	 19,544\\
\hline
Czech (cs)  & 337 		 &	193 &	241 &	268 & 2,810 &	 18,650\\
Danish (da)	& 346    	 &	198	 & 247 &	283	 &  2,951 & 18,340\\
Dutch (nl) & 346   	 &	200 &	247 &	279	 & 2,852 &  17,684\\
\hline

\end{tabular}
\caption{The size of the test sets. The data sets in English are the original test sets.
To evaluate other languages, we translate the data sets from English.}
\label{table:statistic_size}
\end{table*}

\subsection{Evaluation tasks}
We consider two widely used tasks for evaluating word embeddings, the word similarity task and the word analogy task.  
In the word similarity task, each question contains a word pairs and an annotated similarity score. The goal is to predict the similarity score between two words based on the inner product between the corresponding word vectors.
The performance is then measured by the Spearman’s rank correlation coefficient, which estimates the correlation between the model predictions and human annotations. Following the settings in literature, the experiments are conducted on five data sets, \textit{WordSim353}  ~\cite{finkelstein2001placing}, \textit{WordSim Similarity}~\cite{zesch2008using}, \textit{WordSim Relatedness}~\cite{agirre2009study}, \textit{Mechanical Turk} ~\cite{radinsky2011word} and \textit{MEN}~\cite{bruni2012distributional}.

In the word analogy task, we aim at solving analogy puzzles like ``man is to woman as king is to ?'', where the expected answer is ``queen.'' We consider two approaches for generating answers to the puzzles, namely 3CosAdd and 3CosMul (see~\cite{levy2014linguistic} for details). We evaluate the performances on  \textit{Google analogy} dataset ~\cite{Mikolov2013} which contains 8,860 semantic and 10,675 syntactic questions. For the analogy task, only the answer that exactly matches the annotated answer is counted as correct. As a result, the analogy task is more difficult than the similarity task because the evaluation metric is stricter and it requires algorithms to differentiate words with similar meaning and find the right answer. 

To evaluate the performances of models in the low-resource setting, we train word embedding models on Dutch, Danish, Czech and, English data sets collected from Wikipedia. The original Wikipedia corpora in Dutch, Danish, Czech and English contain 216 million, 47 million, 92 million, and 1.8 billion tokens, respectively. To simulate the low-resource setting, we sub-sample the Wikipedia corpora and create a subset of  64 million tokens for Dutch and Czech and a subset of 32 million tokens for English. We will demonstrate how the size of the corpus affects the performance of embedding models in the experiments. 







\begin{table*}[h!]
\centering

\begin{tabular}{c||c|c|c|c|c||c|c}
\hline
Dutch (nl)  & \multicolumn{5}{c||}{Similarity task} & \multicolumn{2}{c}{Analogy task} \\ 
 \hline
Word embedding & WS353 &  Similarity &  Relatedness & M. Turk  & MEN & 3CosAdd & 3CosMul \\ 
 \hline 
GloVe & 35.4 & 35.0 & 41.7 & 44.3  & 11  & 21.2 &	20.2 \\ 
SGNS & 51.9 & 52.9 & 53.5 & $~\textbf{49.8}^{*}$ & 15.4 & 22.1 & 23.6 \\ 
PU-learning & $~\textbf{53.7}^{*}$ & $~\textbf{53.4}^{*}$ & $~\textbf{55.1}^{*}$ & 46.7  & $~\textbf{16.4}^{*}$  & $~\textbf{23.5}^{*}$ & $~\textbf{24.7}^{*}$  \\
\hline
\hline
Danish (da) & \multicolumn{5}{c||}{Similarity task} & \multicolumn{2}{c}{Analogy task} \\ 
 \hline
Word embedding & WS353 &  Similarity &  Relatedness & M. Turk  & MEN & 3CosAdd & 3CosMul \\ 
 \hline 
GloVe & 25.7 &	18.4 & 40.3 & 49.0 & 16.4 & $~\textbf{25.8}^{*}$ & $~\textbf{24.3}^{*}$ \\ 
SGNS & 49.7 & 47.1 & 52.1 & 51.5 & 22.4 & 22.0 & 21.2 \\ 
PU-learning & $~\textbf{53.5}^{*}$ & $~\textbf{49.5}^{*}$ & $~\textbf{59.3}^{*}$ & $~\textbf{51.7}^{*}$ & $~\textbf{22.7}^{*}$ & 22.6 & 22.8  \\
\hline
\hline
Czech (cs) & \multicolumn{5}{c||}{Similarity task} & \multicolumn{2}{c}{Analogy task} \\ 
 \hline
Word embedding & WS353 &  Similarity &  Relatedness & M. Turk  & MEN & 3CosAdd & 3CosMul \\ 
 \hline 
GloVe &  34.3 & 23.2 & 48.9 & 36.5 & 16.2 & 8.9	 & 8.6 \\ 
SGNS & 51.4 & 42.7   & 61.1  & 44.2 &  21.3  & $~\textbf{10.4}^{*}$ & 9.8 \\ 
PU-learning & $~\textbf{54.0}^{*}$ & $~\textbf{45.4}^{*}$ & $~\textbf{65.3}^{*}$ & $~\textbf{46.2}^{*}$ & $~\textbf{21.7}^{*}$ & 9.9 & $~\textbf{10.1}^{*}$  \\
\hline
\hline
English (en) & \multicolumn{5}{c||}{Similarity task} & \multicolumn{2}{c}{Analogy task} \\ 
 \hline
Word embedding & WS353 &  Similarity &  Relatedness & M. Turk  & MEN & 3CosAdd & 3CosMul \\ 
 \hline 
GloVe &  47.9 &	52.1 &	49.5 &	58.8 & 19.1 & 34.3	 & 32.6 \\ 
SGNS & 65.7 & $~\textbf{67.1}^{*}$ & 66.5 & $~\textbf{62.8}^{*}$ & $~\textbf{26.1}^{*}$ &  31.2 & 27.4\\ 
PU-learning & $~\textbf{67.0}^{*}$ & 66.7 & $~\textbf{69.6}^{*}$ & 59.4  & 22.4 & $~\textbf{39.2}^{*}$ & $~\textbf{38.8}^{*}$  \\
\hline
\end{tabular}
\caption{Performance of SGNS, GloVe, and the proposed PU-Learning model in four different languages. Results show that
the proposed PU-Learning model outperforms SGNS and GloVe in most cases when the size of corpus is relatively small (around 50 million tokens). Star indicates it is significant better than  the second best algorithm in the same column according to Wilcoxon signed-rank test. ($p < 0.05$).}
\label{table:comparing2}
\end{table*}

To evaluate the performance of word embeddings in Czech, Danish, and Dutch, we translate the English similarity and analogy test sets to the other languages by using Google Cloud Translation API\footnote{https://cloud.google.com/translate}. However, an English word may be translated to multiple words in another language (e.g., compound nouns). We discard questions containing such words  (see Table \ref{table:statistic_size} for details).
Because all approaches are compared on the same test set for each language, the comparisons are fair.

\subsection{Implementation and Parameter Setting}
We compare the proposed approach with two baseline methods, GloVe and SGNS. 
The implementations of Glove\footnote{https://nlp.stanford.edu/projects/glove} and SGNS\footnote{https://code.google.com/archive/p/word2vec/} and provided by the original authors, and we apply the default settings when appropriate. 
The proposed PU-Learning framework is implemented based on \newcite{Yu2016}. With the implementation of efficient update rules, our model requires less than 500 seconds to perform one iteration over the entire text8 corpus, which consists of 17 million tokens~\footnote{\url{http://mattmahoney.net/dc/text8.zip}}.  All the models are implemented in C++.

We follow~\newcite{Levy2015}\footnote{https://bitbucket.org/omerlevy/hyperwords} to set windows size as 15, minimal count as 5, and dimension of word vectors as 300 in the experiments.
Training word embedding models involves selecting several hyper-parameters. However, as the word embeddings are usually evaluated in an unsupervised setting (i.e., the evaluation data sets are not seen during the training), the parameters should not be tuned on each dataset. To conduct a fair comparison, we tune hyper-parameters on the text8 dataset. For GloVe model, we tune the discount parameters $x_{max}$ and find that $x_{max} = 10$ performs the best. SGNS has a natural parameter $k$ which denotes the number of negative samples. Same as \newcite{Levy2015}, we found that setting $k$ to 5 leads to the best performance. For the PU-learning model, $\rho$ and $\lambda$ are two important parameters that denote the unified weight of zero entries and the weight of regularization terms, respectively. We tune  $\rho$  in a range from $2^{-1}$ to $2^{-14}$ and $\lambda$ in a range from $2^{0}$ to $2^{-10}$. We analyze the sensitivity of the model to these hyper-parameters in the experimental result section. 
The best performance of each model on the text8 dataset is shown in the Table \ref{table:comparingontext8}. It shows that PU-learning model outperforms two baseline models. 

\section{Experimental Results}
\label{sec:exp-para}

\begin{figure*}[!h]
\vspace{10pt} 

\includegraphics[width=1\linewidth]{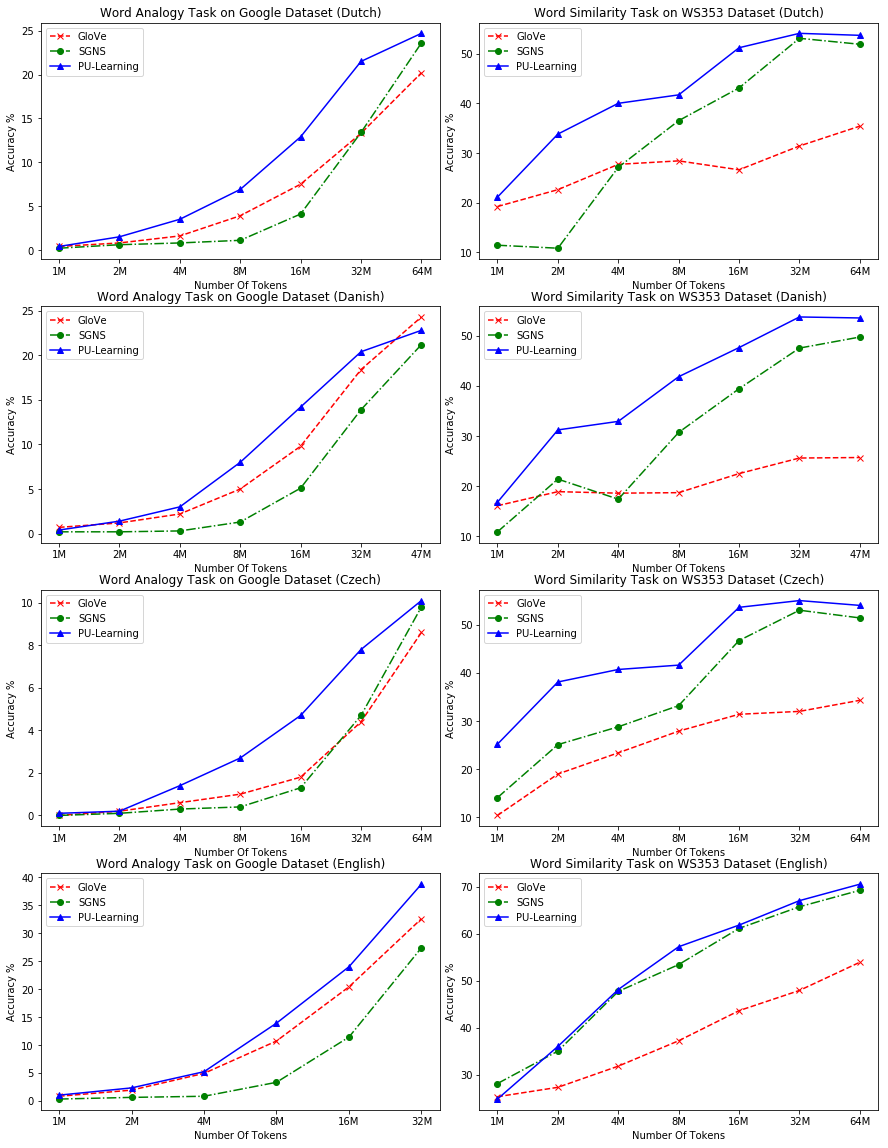}

\caption{Performance change as the corpus size growing  (a)  on the Google word analogy task (on the left-hand side) and (b) on the WS353 word similarity task (on the right-hand side). We demonstrate the performance on four languages, Dutch, Danish, Czech and English datasets. Results show that PU-Learning model consistently outperforms SGNS and GloVe when the size of corpus is small.}
\label{fig:comparing}
\end{figure*}

\begin{figure*}[!htp]
\vspace{10pt} 

\includegraphics[width=1\linewidth]{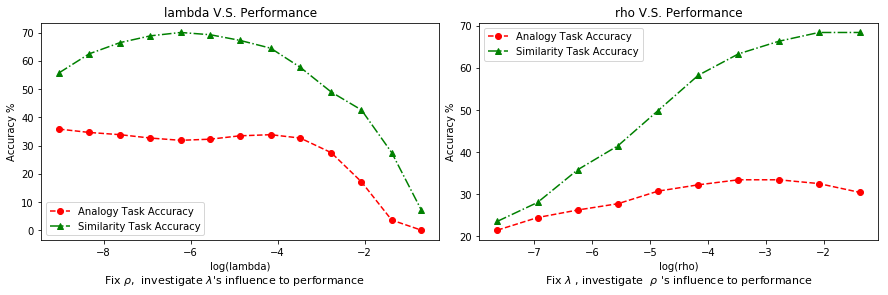}

\caption{Impact of $\rho$ and $\lambda$ in the PU-Learning framework.}
\label{fig:comparing_2_in_1}
\end{figure*}

We compared the proposed PU-Learning framework with two popular word embedding models -- 
SGNS~\cite{Mikolov2013SGNS} and Glove~\cite{pennington2014glove} on English and three other languages.  The experimental results are reported in Table~\ref{table:comparing2}.
The results show that the proposed PU-Learning framework outperforms the two baseline approaches significantly in most datasets. This results confirm that the unobserved word pairs carry important information and the PU-Learning model leverages such information and achieves better performance. To better understand the model, we conduct detailed analysis as follows.




\paragraph{Performance v.s. Corpus size}
We investigate the performance of our algorithm with respect to different corpus size, and plot the results in Figure~\ref{fig:comparing}. The results in analogy task are obtained by 3CosMul method~\cite{levy2014linguistic}. As the corpus size grows, the performance of all models improves, and the PU-learning model consistently outperforms other methods in all the tasks. However, with the size of the corpus increases, the difference becomes smaller. This is reasonable as when the corpus size increases the number of non-zero terms becomes smaller and the PU-learning approach is  resemblance to Glove.








\paragraph{Impacts of $\rho$ and $\lambda$}
We investigate how sensitive the model is to the hyper-parameters, $\rho$ and $\lambda$. Figure~\ref{fig:comparing_2_in_1} shows the performance along with various values of $\lambda$ and $\rho$ when training on the text8 corpus, respectively. Note that the x-axis is in $\log$ scale.
When $\rho$ is fixed, a big $\lambda$ degrades the performance of the model significantly. This is because when $\lambda$ is too big the model suffers from under-fitting. The model is less sensitive when $\lambda$ is small and in general, $\lambda = 2^{-11}$ achieves consistently good performance.

When $\lambda$ is fixed, we observe that large $\rho$ (e.g., $\rho \approx 2^{-4}$) leads to better performance. As $\rho$ represents the weight assigned to the unobserved term, this result confirms that the model benefits from using the zero terms in the co-occurrences matrix. 

\section{Conclusion}
In this paper, we presented a PU-Learning framework for learning word embeddings of low-resource languages. We evaluated the proposed approach on English and other three languages and showed that the proposed approach outperforms other baselines by effectively leveraging the information from unobserved word pairs.

In the future, we would like to conduct experiments on other languages where available text corpora are relatively hard to obtain. We are also interested in applying the proposed approach to domains, such as legal documents and clinical notes, where the amount of accessible data is small. Besides, we plan to study how to leverage other information to facilitate the training of word embeddings under the low-resource setting.

\section*{Acknowledge}
This work was supported in part by National Science Foundation Grant IIS-1760523, IIS-1719097 and an NVIDIA Hardware Grant.

\bibliography{naaclhlt2018}
\bibliographystyle{acl_natbib}


\end{document}